# A Novel Robust Method to Add Watermarks to Bitmap Images by Fading Technique

Firas A. Jassim

*Abstract*—Digital water marking is one of the essential fields in image security and copyright protection. The proposed technique in this paper was based on the principle of protecting images by hide an invisible watermark in the image. The technique starts with merging the cover image and the watermark image with suitable ratios, i.e., 99% from the cover image will be merged with 1% from the watermark image. Technically, the fading process is irreversible but with the proposed technique, the probability to reconstruct the original watermark image is great. There is no perceptible difference between the original and watermarked image by human eye. The experimental results show that the proposed technique proven its ability to hide images that have the same size of the cover image. Three performance measures were implemented to support the proposed techniques which are MSE, PSNR, and SSIM. Fortunately, all the three measures have excellent values.

## I. INTRODUCTION

Rapid growth of digital multimedia technologies causes an easy copying and distribution of digital multimedia.

Hence, the copy prevention and copyright protection of digital multimedia has become a significant matter. The most widely technique to correct this situation is digital watermarking. Digital watermarking is a method for embedding information into an image which can be extracted lately. Actually, digital watermarking was invented for variety of goals including identification, copy prevention and authentication purposes [4]. Watermarking is used to protect the copyright of the original owners, to make their claim stronger, should the image be used without permission by someone else [20]. At the beginning, it must be differentiated between watermarking and steganography. Actually, both the two methods were implemented in the same filed which the security field but with different approaches. Digital watermarking is changing an image in a way such that you can see some text or background image without actually corrupting the image. In digital watermarking, the focus is on ensuring that nobody can remove or alter the content of the watermarked data, even though it might be plainly obvious that it exists. Steganography on the other hand, focuses on making it extremely difficult to tell that a secret message exists at all [16]. Theoretically, digital watermarking is considered as a special case of information hiding. Digital watermarking is the process of embedding information into digital multimedia content such that the information (the watermark) can later be extracted or detected for a variety of purposes including copy prevention and control. Digital watermarking has become an active and important area of research, and development and commercialization of watermarking techniques is being considered essential to help address some of the challenges faced by the rapid growth of digital content. The key difference between information hiding and watermarking is the absence of an active adversary. In watermarking applications like copyright protection and authentication, there is an active adversary that would attempt to remove, invalidate or forge watermarks. In information hiding there is no such active adversary as there is no value associated with the act of removing the information hidden in the content. Nevertheless, information hiding techniques need to be robust against accidental distortions [14].

Steganography is the act of adding a hidden message to an image or other media file. It is similar to encrypting a document, but instead of running it through a cipher, the document is broken up and stored in unused, or unnoticeable, bits within the overall image. Watermarking is similar, but has a completely different purpose. Placing a watermark in an image or other media file serves to identify the artist or author of the work. It isn't so much an attempt to hide a message as it is to tag a document for later identification [21].

The organization of this paper is as follows: In section II, a brief background concerning digital watermarking and its basic schemes implemented. The proposed fading technique for robust image watermarking has been presented in section III. The experimental results and simulation outcomes have been discussed in section IV. Finally, the main conclusions that come from this paper were presented in section V.

## II. WATERMARKING PRELIMINARIES

Watermarking schemes can be robust or fragile. Robust watermarks are designed to resist to malicious or intentional distortions, such as general image processing and geometric distortions. On the contrary, fragile watermarks are required for the purpose of authentication and verification [12]. Watermarking is of two types; visible watermarking and invisible watermarking. Concerning visible Watermarking, as

---

F. A. Author is with the Faculty of Multimistrative Sciences, Management Information Systems Department, Irbid National University, Irbid 2600, Jordan, (Email: firasajil@yahoo.com)

the name suggests, it refers to the information visible on the image or video or picture. Visible watermarks are typically logos or text. For example, in a TV broadcast, the logo of the broadcaster is visible at the right side of the screen. On the other hand, invisible watermarking refers to adding information in a video or picture or audio as digital information. It is not visible or ocular, but it can be detected by different approaches. It may also be a form or type of steganography and is used for widespread use. It can be retrieved easily. The essential applications of watermarking are copyright protection, source tracing, and annotation of photographs.

There are a variety of schemes for embedding the watermark into the original image [2], [8]. Typical schemes for digital watermarking were based on transform-domain techniques with discrete cosine transform (DCT), discrete wavelet transform (DWT) [13]. AS part of wavelet transform, Haar wavelet transform for copyright protection was proposed as a robust-invisible watermarking technique [17]. A genetic watermarking approach based on transform-domain technique was discussed by [3]. Moreover, a robust digital image watermarking method using dual tree complex wavelet transform was presented by [19]. The implementation of principle component analysis (PCA) into digital video watermarking was discussed by [9].

### III. PROPOSED FADING TECHNIQUE

As mentioned previously, the proposed technique is based on the fading principle between two images. The cover image and the logo image are faded with different fading ratios. Actually, it is an essential problem to determine the correct fading ratios for both the cover and logo images. First of all, according to [7], the general formula for fading two images is as follows:

$$h = \alpha_1 f + \alpha_2 g \qquad (1)$$

where $h$, $f$, and $g$ are the representations for Watermarked, Original and Logo images, respectively. Moreover, $\alpha_1$ and $\alpha_2$ are both the fading ratios that satisfy the following equation:

$$\alpha_1 + \alpha_2 = 1 \qquad (2)$$

Practically, changing $\alpha_1$ and $\alpha_2$ may change the ocular differences in fading between the original and Logo images. In accordance with Fig. 1, different ratios have been implemented to show the overlapping between the original image (Lena.bmp) and the Logo image (Logo.bmp).

It can be seen that, there is a positive relationship between the clarity of the image and the fading ratios (either $\alpha_1$ or $\alpha_2$). Hence, increasing $\alpha_1$ and decreasing $\alpha_2$ will highly makes the original image appear more clearly and the logo image vanishes progressively. Generally, the proposed technique can be summarized in Fig. 2.

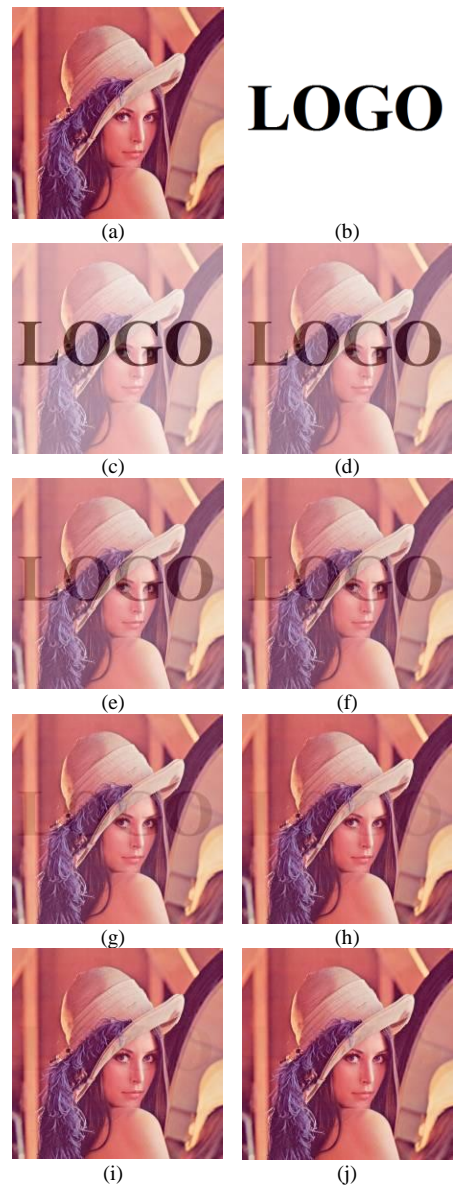

Fig. 1. (a) Original image (b) Logo image (c) 50%, 50% (d) 60%, 40% (e) 75%, 25% (f) 80%, 20% (g) 90%, 10% (h) 95%, 5% (i) 97%, 3% (j) 99%, 1%, ratios are for original and logo images, respectively

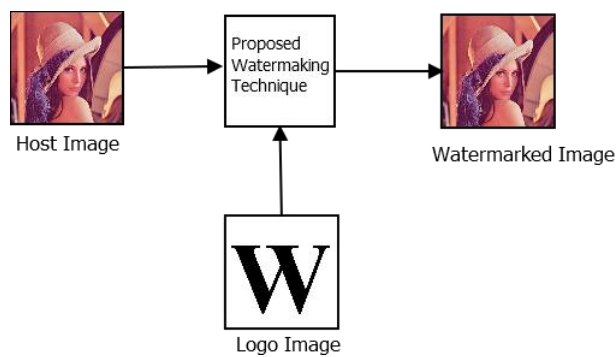

Fig. 2. Watermarking process

*A. Extracting Watermark*

In order to extract the logo form the watermarked image, a general formula has been derived to retrieve logo image (g). According to equation (1), one can solve it for g as follows:

$$g = Round\left(\frac{h - Round(\alpha_1 f)}{\alpha_2}\right) \quad (3)$$

*B. Robustness against Attacks*

The fundamental reason of watermarking invention was to protect media copyright in the first place from outside attack. The quality of watermarking algorithm depends on its ability to survive against various kinds of attacks that try to remove or destroy the watermark. However, attempting to remove or destroy the watermark should produce a noticeable debility in image quality. The typical attacks are linear filtering (e.g., low-pass and high-pass filtering), nonlinear filtering (e.g., median filtering), geometric distortions (e.g., scaling, rotation, and cropping), compression, and blurring, etc [4]. Moreover, the basic requirements for adaptive image watermarking schemes are transparency, capacity, and robustness [5]. Firstly, transparency means that the watermark is not visible in the image under typical viewing conditions. Secondly, the capacity, that represents the maximum amount of information that can be hidden and retrieved successfully. Thirdly, robustness means that the ability of the watermarking technique to survive the attempts of removing the hidden information [1].

## IV. EXPERIMENTAL RESULTS

In this section, both a practical and visual evidences have been implemented to support the proposed technique. Therefore, a variety of test images have been tested (figure 3) against different logo images (figure 4) to construct a watermarked image. According to figure (5), the ocular results are highly acceptable. Furthermore, three performance evaluations have been accomplished to sustain the proposed technique. The first performance measure is the Mean square Error (MSE) [7]. MSE is the cumulative squared error between the watermarked and the original images which can be represented as:

$$MSE = \frac{1}{NM} \sum_{x=1}^{N} \sum_{y=1}^{M} [f(x,y) - g(x,y)]^2 \quad (4)$$

The second measure is the widely implemented measure in most image processing fields and that is the Peak-Signal-to-Noise Ratio (PSNR) [11][7]. Actually, it produces better results than MSE because it takes into consideration the peaks of the signal. The PSNR block computes the peak signal-to-noise ratio, in decibels, between two images. This ratio is often used as a quality measurement between the original and a compressed image. The higher the PSNR, the better the quality of the watermarked image. The general formula of the PSNR can be expressed as:

$$PSNR = 10 \cdot \log_{10}\left(\frac{255^2}{\sqrt{MSE}}\right) \quad (5)$$

The final performance measure is the structure similarity index (SSIM) which was proposed by [23]. The mathematical representation for the SSIM is as follows:

$$SSIM(x,y) = \frac{(2\mu_x\mu_y + C_1)(2\sigma_{xy} + C_2)}{(\mu_x^2 + \mu_y^2 + C_1)(\sigma_x^2 + \sigma_y^2 + C_2)} \quad (6)$$

The dynamic range of SSIM is [−1, 1]. The best value of SSIM is 1 which is achieved if and only if x and y are identical. Nearly tabulate regions, the denominator of the contrast comparison formula is approach to zero, which makes the algorithm unstable. Therefore, inserting the small constants $C_1$ and $C_2$ [23].

The whole three performance measures have been shown in Table I for both the watermarked and extracted logo images. The numerical values for the three measures are very satisfactory and persuasive.

In order to test the robustness of the proposed algorithm, a number of image processing attacks were applied to the watermarked image. As mentioned previously, the most widely implemented attacks are geometric distortion (scaling, rotation, Gaussian white noise, and salt & peppers noise), compression, linear filtering (low and high pass), non-linear filtering (median filter, wiener filter), and blurring, Fig. 6.

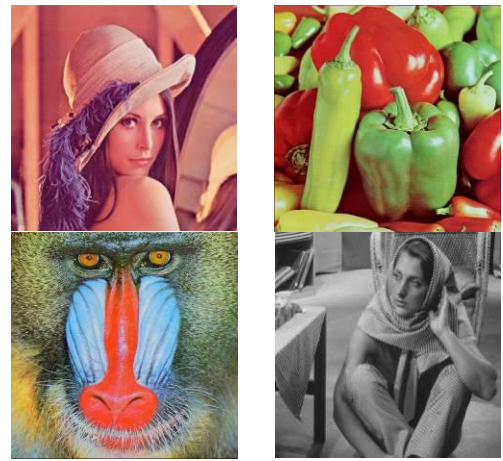

Fig. 3. Test images

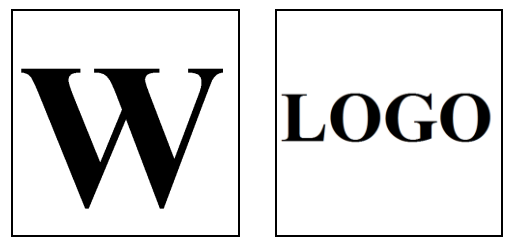

Fig. 4. Watermarks images Logow.bmp and Logo1.bmp, respectively

Table I: MSE, PSNR, and SSIM for test images

|  | MSE | PSNR | SSIM |
|---|---|---|---|
| Lena(with Logo1) | 2.4334 | 44.2687 | 0.9978 |
| Lena(with LogoW) | 2.2602 | 44.5894 | 0.9978 |
| Logo1(from Lena) | 0.9022 | 48.5780 | 0.9998 |
| LogoW(from Lena) | 0 | ∞ | 1.0000 |
| Baboon(with Logo1) | 2.1951 | 44.7163 | 0.9993 |
| Baboon(with LogoW) | 2.1141 | 44.8795 | 0.9993 |
| Logo1(from Baboon) | 0.9022 | 48.5780 | 0.9998 |
| LogoW(from Baboon) | 0 | ∞ | 1.0000 |
| Peppers(with Logo1) | 2.5741 | 44.0246 | 0.9976 |
| Peppers(with LogoW) | 2.5512 | 44.0633 | 0.9975 |
| Logo1(from Peppers) | 0.9022 | 48.5780 | 0.9998 |
| LogoW(from Peppers) | 0 | ∞ | 1.0000 |
| Barbara(with Logo1) | 2.7903 | 43.6742 | 0.9990 |
| Barbara(with LogoW) | 2.6065 | 43.9702 | 0.9990 |
| Logo1(with Barbara) | 0.9022 | 48.5780 | 0.9998 |
| LogoW(with Barbara) | 0 | ∞ | 1.0000 |

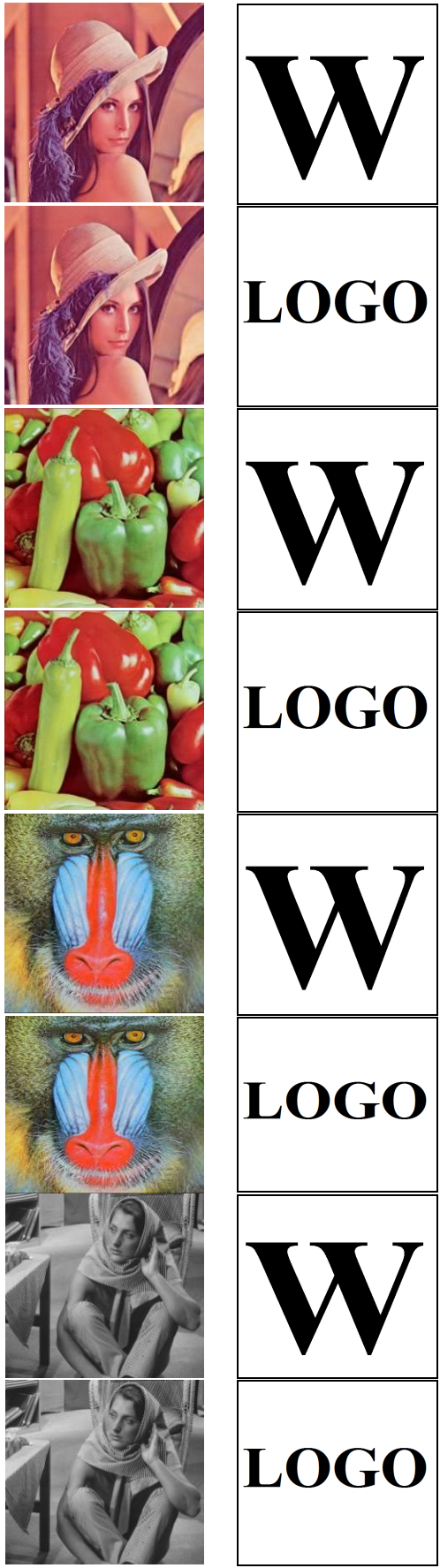

Fig. 5. Watermarked image (right) Extracted watermark (left)

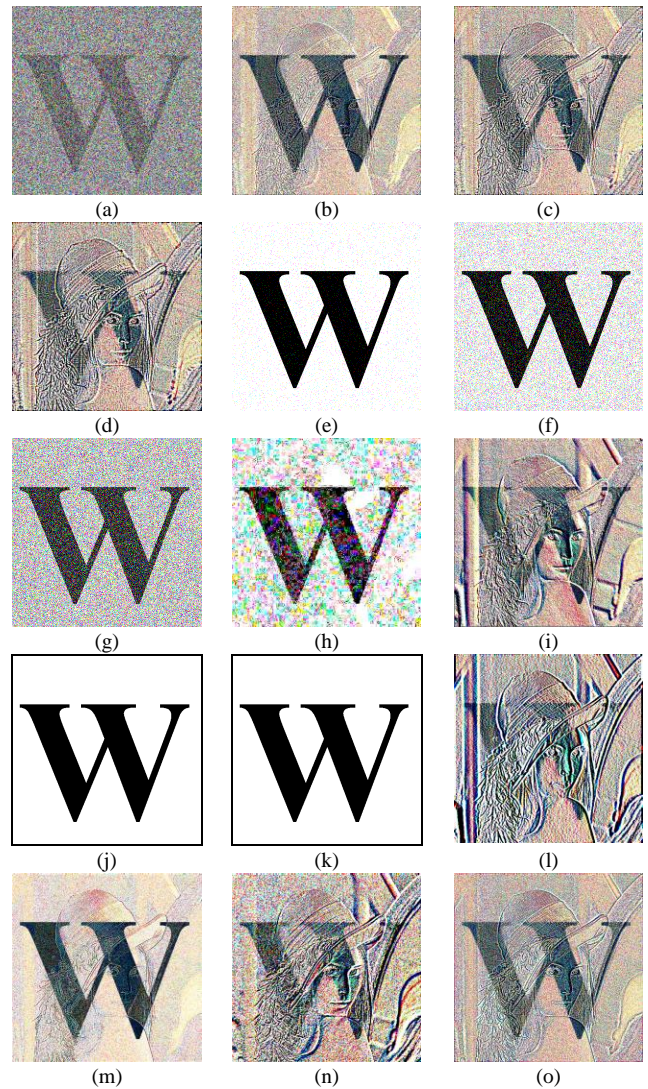

Fig. 6. (a) Gauss (Mean=0, Var=0.001) (b) Median (3X3) (c) Median(5X5) (d) Median(7X7) (e) (Salt & Peppers) 2% (f) (Salt & Peppers) 20% (g) (Salt & Peppers) 50% (h) JPEG (i) Rotate (45) (j) Rotate (90) (k) Rotate (180) (l) Blurring (m) Scale (1536X1536) (3X) (n) Scale (256X256)(0.5X) (o) Wiener Filter

## V. CONCLUSION

In this paper, a novel robust image watermarking technique was proposed. The essential conclusion that comes from the proposed technique is the high robustness to (almost) attacks that may be implemented by media forgers. In most of the attacks, the hidden watermark could be always extracted either complete or incomplete, i.e. there is always a recognizable watermark. The proposed technique may be embedded into image, video, or audio. Another important conclusion is that, the proposed fading technique produces an exact (100%) extracted watermark when rotation with (45, 90, 180, 270, 360) degrees. As a future step, the proposed technique could be embedded into video and audio watermarking. According to the experimental results and high error metrics, the novel proposed fading technique proven that it is very simple and robust against multiple attacks. Furthermore, the novel technique is the first of its type that embed a watermark that has the same dimensions with the original cover image. It must be mentioned that the payload of the proposed technique is high, i.e. the embedded watermark does not affect the size (in Kilobytes) of the cover image.

**First A. Jassim** was born in Baghdad, Iraq, in 1974. He received the B.S. and M.S. degrees in Applied Mathematics and Computer Applications from Al-Nahrain University, Baghdad, Iraq, in 1997 and 1999, respectively, and the Ph.D. degree in Computer Information Systems (CIS) from the Arab University for Banking and Financial Sciences, Amman, Jordan, in 2012. In 2012, he joined the faculty of Business Administration, Management Information Systems Department, Irbid National University, Irbid, Jordan, where he is currently an assistance professor. His current research interests are image compression, image interpolation, image segmentation, image enhancement, steganography, and simulation.